\documentclass[conference]{IEEEtran}
\IEEEoverridecommandlockouts
\usepackage{cite}
\usepackage{amsmath,amssymb,amsfonts}
\usepackage{algorithmic}
\usepackage{graphicx}
\usepackage{textcomp}
\usepackage{xcolor}
\usepackage{multirow}
\usepackage{mathtools}
\usepackage[
singlelinecheck=false 
]{caption}
\def\BibTeX{{\rm B\kern-.05em{\sc i\kern-.025em b}\kern-.08em
    T\kern-.1667em\lower.7ex\hbox{E}\kern-.125emX}}
\begin{document}

\title{Dynamic Curvature-Constrained Path Planning}

\author{\IEEEauthorblockN{Nishkal Gupta Myadam}
\IEEEauthorblockA{
\textit{mnishkalgupta@gmail.com}\\
Hyderabad, India \\}

}

\maketitle

\begin{abstract}
Effective path planning is a pivotal challenge across various domains, from robotics to logistics and beyond. This research is centered on the development and evaluation of the Dynamic Curvature-Constrained Path Planning Algorithm (DCCPPA) within two-dimensional space. DCCPPA is designed to navigate constrained environments, optimizing path solutions while accommodating curvature constraints.The study goes beyond algorithm development and conducts a comparative analysis with two established path planning methodologies: Rapidly Exploring Random Trees (RRT) and Probabilistic Roadmaps (PRM). These comparisons provide insights into the performance and adaptability of path planning algorithms across a range of applications.This research underscores the versatility of DCCPPA as a path planning algorithm tailored for 2D space, demonstrating its potential for addressing real-world path planning challenges across various domains.
\end{abstract}


\section{Introduction}
Path planning, a fundamental challenge in the field of computer science and robotics, serves as a cornerstone for the successful operation of autonomous systems in various domains. The ability to determine optimal and collision-free routes from a starting point to a predefined goal, while considering the complexities of the environment, is a critical component in applications ranging from autonomous vehicles and drones to industrial automation and beyond. As the capabilities and demands of these systems continue to evolve, so too must the methodologies employed to address the path planning problem.

This research embarks on a multifaceted exploration, with a primary focus on the development and analysis of the Dynamic Curvature-Constrained Path Planning Algorithm (DCCPPA). DCCPPA offers a novel approach tailored for two-dimensional (2D) space, marked by its capacity to maneuver through constrained environments, optimizing trajectories while accommodating curvature constraints.

The journey through this research, however, extends beyond the singular pursuit of algorithmic innovation. To offer a holistic understanding of the DCCPPA's strengths and adaptability, we embark on a comparative voyage that introduces two stalwarts of path planning: Rapidly Exploring Random Trees (RRT) and Probabilistic Roadmaps (PRM)\cite{b1}. These renowned algorithms, celebrated for their versatility and application across diverse domains, serve as benchmarks for evaluating the performance of DCCPPA.

The synergy between theory, practical implementation, and empirical analysis forms the core of our research methodology. This endeavor encompasses a comprehensive literature review, delving into the foundations of these algorithms, and their practical applications. Through diligent algorithm development and rigorous evaluation, we aim to provide the reader with a comprehensive understanding of the relative merits and adaptability of DCCPPA, RRT, and PRM.

While the roots of our investigation lie in path planning, the implications and applications of our findings transcend the boundaries of any single domain. DCCPPA, designed for 2D space, offers insights that reach beyond robotics, with potential applications in domains as diverse as autonomous logistics and video game design.

In the sections that follow, we will delve into the nuances of each algorithm, explore their practical implementation, present the results of our comparative analysis, and conclude with a reflection on the implications of our research. By the end of this journey, the reader will not only grasp the intricate world of path planning but also the universality of DCCPPA's contributions to the broader landscape of algorithmic innovation.

\section{Research Methodology}

\subsection{Research Methodology}
A hybrid research methodology is adapted for this paper, where both literature review and formal experiment is conducted in order to attain the objective of designing a more robust path planning algorithm called DCCPPA along with it's comparison with RRT and PRM. Below image depicts a deatiled workflow of the research methodology.

\begin{figure}[h!]
    \centering
    \includegraphics[scale=0.3]{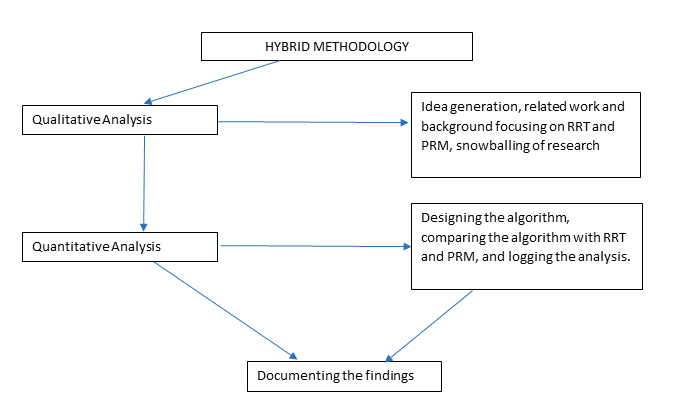}
    \caption{Flow Chart of Research Methodology}
    \label{fig:my_label}
\end{figure}

\subsection{Related Work}
This subsection provides an in-depth exploration of the theoretical foundations and practical applications of key path planning algorithms.
\subsubsection{Probabilistic Road Map}
Probabilistic Roadmaps (PRM) constitute a foundational approach to path planning, renowned for their versatility and robustness in handling high-dimensional configuration spaces. Developed by Kavraki et al. \cite{b1} in the late 1990s, PRMs operate on the principle of probabilistic sampling, constructing a roadmap of feasible paths through random configurations . The roadmap is then used to efficiently find paths between start and goal configurations, making PRMs applicable to a wide range of robotic systems.

Applications of PRMs span various domains, including robotics, manufacturing, and motion planning for autonomous vehicles. Their ability to handle complex, high-dimensional spaces and adapt to dynamic environments has solidified PRMs as a go-to choice in scenarios where deterministic approaches may falter.

\begin{table}[h!]

\caption{Advantages and Disadvantages of PRM}

\begin{tabular}{|p{1.3in}|p{1.3in}|}

\hline

Advantages & Disadvantages\\\hline 
Versatility: PRMs are versatile and applicable to a broad range of robotic systems and environments. & Computational Cost: The construction of the road map can be computationally expensive, especially in high-dimensional spaces.\\\hline
Handling High Dimensions: PRMs excel in handling high-dimensional configuration spaces, making them suitable for complex scenarios. & Memory Requirements: PRMs may require substantial memory for storing the roadmap, which can be a constraint in resource-limited systems. \\\hline
Adaptability: They can adapt to dynamic environments by updating the roadmap based on changes in the workspace. & Initial Roadmap Generation: Generating an initial roadmap may be challenging, particularly in environments with intricate obstacle layouts. \\\hline

\end{tabular}
\hfill\break
    
\end{table}

\subsubsection{Rapidly Exploring Random Trees (RRT):}
RRT represent a paradigm shift in path planning algorithms, emphasizing the rapid exploration of configuration spaces. Developed by LaValle and Kuffner in the early 2000s, RRTs excel in scenarios where a quick and adaptive search is crucial. The algorithm incrementally grows a tree structure by exploring the space randomly, biased toward unexplored regions, making it particularly effective in dynamic environments\cite{b2}. This randomness is a deliberate and essential feature of RRT, allowing it to efficiently explore vast and complex spaces.

\begin{table}[h!]

\caption{Advantages and Disadvantages of RRT}

\begin{tabular}{|p{1.3in}|p{1.3in}|}

\hline

Advantages & Disadvantages\\\hline 
Rapid Exploration: RRTs rapidly explore configuration spaces, making them suitable for dynamic environments. & Lack of Determinism: The randomness in exploration may lead to non-deterministic behavior.\\\hline
Incremental Growth: The tree structure grows incrementally, allowing adaptability to changes in the environment during runtime. & Suboptimal Paths: RRTs might produce suboptimal paths, and their efficiency depends on the sampling strategy. \\\hline
Applicability to Non-holonomic Systems: RRTs are well-suited for systems with non-holonomic constraints. & Sensitivity to Parameters: The performance of RRTs can be sensitive to parameters like step size and sampling strategy. \\\hline

\end{tabular}
\hfill\break
    
\end{table}

RRTs have found widespread use in robotics, particularly in scenarios with non-holonomic constraints and complex obstacle layouts. Their adaptability and efficiency in high-dimensional spaces have positioned RRTs as a staple in the toolbox of path planning algorithms.
\hfill\break
\hfill\break
For the scope of this research paper, our focus has been directed towards Probabilistic Roadmaps (PRM) and Rapidly Exploring Random Trees (RRT), two prominent algorithms in the field of path planning. This decision is driven by the overarching goal of developing a novel algorithm Dynamic Curvature-Constrained Path Planning Algorithm (DCCPPA)—specifically tailored for 2D environments with curvature constraints.
While PRM and RRT represent key players in the realm of path planning, it's imperative to acknowledge the diversity of algorithms employed across various applications. Noteworthy alternatives include A* (A Star), known for its optimality guarantees, and others such as Dijkstra's algorithm and Genetic Algorithms\cite{b4}. 

In the following section, we will unravel the intricacies of DCCPPA, detailing its design, implementation, and performance in comparison to these established methodologies.
\section{Dynamic Curvature Constrained Path Planning Algorithm (DCCPPA)}
Path planning in 2D environments poses unique challenges, especially when considering constraints related to curvature. The existing algorithms, while powerful, may not always efficiently handle scenarios where navigating through varying curvatures is crucial. In response to this gap, we introduce the Dynamic Curvature-Constrained Path Planning Algorithm (DCCPPA), a novel approach specifically tailored for 2D spaces.

\textbf{Motivation:}

The motivation behind the development of DCCPPA stems from the need for a path planning algorithm that not only navigates efficiently through a 2D environment but also addresses curvature constraints. Existing algorithms like PRM and RRT, while effective in many scenarios, may struggle when confronted with environments where the optimal path involves trajectories with varying curvature \cite{b3}.

\textbf{Reason behind developing new algorithm:}
\begin{itemize}
    \item \textbf{Curvature-Centric Design:} DCCPPA is conceived with curvature constraints at its core. By prioritizing the efficient traversal of environments with curvature variations, the algorithm aims to outperform existing methods in scenarios where these constraints significantly impact the optimal path.
    \item \textbf{Reduced Sampling Complexity:}  One key objective is to streamline the path planning process by minimizing the number of sampling nodes. DCCPPA achieves this by strategically placing nodes along the perimeter of obstacles, ensuring a more efficient exploration of the configuration space.
\end{itemize}

\subsection{Architecture Design}
\begin{itemize}
    
\item{Node Representation:}
DCCPPA employs a node-based representation for the configuration space. Each node encapsulates information about the robot's position and orientation.
\item{Graph Structure:}
The algorithm constructs a graph that captures the connectivity of the configuration space. Edges represent feasible paths between nodes, and the graph evolves dynamically during the exploration process.
\item{Sampling Strategy:}
DCCPPA strategically places sampling nodes along the perimeters of obstacles to ensure thorough exploration. The algorithm optimizes the density of sampling based on curvature constraints and obstacle characteristics.
\item{Adaptive Exploration:}
To enhance adaptability, DCCPPA incorporates mechanisms for adaptive exploration. The algorithm dynamically adjusts its sampling strategy based on the local curvature requirements, allowing for efficient exploration of complex environments.
\end{itemize}

\subsection{Mathematical Formulation}
\subsubsection{Objective Function}
In DCCPPA, the objective function J(p) combines two important aspects:
\begin{itemize}
    \item Path Length Component: The term Path Length (p) represents the length or distance of the path p from the starting point to the goal. Minimizing this component encourages the algorithm to find shorter paths, which is often desirable in practical applications to reduce travel time or energy consumption.
    \item Curvature Deviation Component: Curvature Deviation (p) introduces a curvature-related criterion. It penalizes paths that deviate significantly from specified curvature constraints. The beta parameter allows you to control the influence of curvature in the overall optimization. This is particularly relevant when navigating around obstacles where certain curvature constraints are preferred.
\end{itemize}

The objective function aims to strike a balance between finding paths that are short and paths that adhere to specified curvature constraints. 

\[
J(p) = \text{{Path Length}}(p) + \beta \cdot \text{{Curvature Deviation}}(p)
\]

\text{{where,}}

\[
\text{{Path Length}}(p) = \sum_{i=1}^{N-1} \| p_i - p_{i+1} \|
\]

\[
\text{{Curvature Deviation}}(p) = \sum_{i=1}^{N} \left( \frac{1}{r_i + d_i} \right)
\]

\text{{Here,}}
\begin{align*}
& N \text{{ is the total number of points in the path,}} \\
& p_i \text{{ represents the }} i\text{{-th point in the path,}} \\
& r_i \text{{ is the radius of the }} i\text{{-th obstacle,}} \\
& d_i \text{ is the Euclidean distance from the } i\text{-th point in the path} \\
& \quad \text{to the center of the } i\text{-th obstacle, and} \\
& \beta \text{ is the weighting factor for balancing path length} \\
&   \quad \text{and curvature deviation.}
\end{align*}

\subsubsection{Curvature Constraints}

DCCPPA evaluates the curvature $\kappa$ at each point along the path, ensuring that $\kappa \leq \text{{Curvature Threshold}}$:

\[
\kappa = \sum_{i=1}^{N} \frac{1}{r_i + d_i}
\]

where:
\begin{align*}
& r_i \text{ is the radius of the $i$-th obstacle,} \\
& d_i \text{ is the Euclidean distance from the point to the center of the} \\
& \quad i\text{-th obstacle,} \\
& N \text{ is the number of obstacles.}
\end{align*}

\subsubsection{Local Search}
DCCPPA employs a local search strategy to optimize the path. Given a current point $\mathbf{P_{\text{current}}}$, the next point $\mathbf{P_{\text{next}}}$ is determined using a simulated gradient descent:

\[
\mathbf{P_{\text{next}}}(\mathbf{x}, \mathbf{y}) = \mathbf{P_{\text{current}}}(\mathbf{x}, \mathbf{y}) + \text{{step size}} \cdot \frac{(\mathbf{P_{\text{goal}}} - \mathbf{P_{\text{current}}})}{\|\mathbf{P_{\text{goal}}} - \mathbf{P_{\text{current}}}\|}
\]

The step size is constrained to a maximum value. $\mathbf{P_{\text{goal}}}$ is the goal point.
\hfill\break
\subsubsection{Global Search}

DCCPPA performs a global search by randomly sampling points in the configuration space until a point $\mathbf{P_{\text{next}}}$ is found that satisfies the curvature constraints.

\hfill\break
\subsubsection{Path Planning}
The path planning process involves iteratively selecting between local and global search strategies based on the current curvature conditions. The algorithm continues this process until the distance between the current point and the goal is below a specified threshold.

\subsubsection{Collision Detection and Avoidance}
Collision detection and Avoidance is incorporated into the algorithm as a safe check

\subsection{How is DCCPPA different?}
DCCPPA introduces a unique combination of global and local searches to achieve effective path planning. In its global search, random sampling is employed to explore diverse regions of the configuration space. However, unlike traditional random sampling approaches, DCCPPA integrates a directional curvature constraint. This constraint guides the global search, injecting a sense of directionality and purpose into the random sampling process.

\section{Results}
This section is a critical component of this research paper, shedding light on the empirical performance and efficiency of the DCCPPA algorithm in comparison to established path planning algorithms, namely PRM and RRT. Through a systematic evaluation, this section aims to provide insights into the effectiveness of DCCPPA across various scenarios, considering both its standalone performance and comparative analysis.
\hfill\break
\textbf{NOTE:} 
\hfill\break
\textit{It is essential to acknowledge the inherent stochastic nature of the path planning algorithms employed in this study—RRT (Rapidly-exploring Random Trees), PRM (Probabilistic Roadmaps), and DCCPPA (Directional Curvature Constrained Path Planning Algorithm). Due to the random sampling strategies inherent in RRT and PRM, the exact paths generated by these algorithms may exhibit slight variations in each run. This randomness arises from the algorithms' adaptive exploration of the configuration space, contributing to their ability to navigate complex environments. Consequently, the number of steps, representing the path length, may show variability between runs. In the case of DCCPPA, which integrates a directional curvature constraint in its global search, the randomness is more purposeful. Despite this, the number of steps might still exhibit slight fluctuations due to the local and global search strategies. Our focus in comparing these algorithms is not merely on the absolute number of steps but on discerning the comparative trends and efficiencies. We aim to assess how consistently each algorithm finds paths under varying conditions and, particularly, how DCCPPA, with its unique directional exploration, tends to demonstrate reduced step counts in our experiments.Through statistical analysis and repeated experiments, we present an overview of the comparative performances, highlighting the strengths and tendencies of each algorithm in terms of path planning efficiency and adaptability.}

\subsection{Experiment Setup}
This section gives us the information regarding the requirements for the experiment, scenarios as well the metrics we are going to use for evaluating the performances of the algorithms. 

\subsubsection{Requirements}
Have a stable operating system with python version 3.8 installed in it, along with the git repository cloned into your environment where you would like to conduct the experiment inside your system. 
\hfill\break
Steps to follow for running the experiment:
\begin{itemize}
    \item git clone \textit{https://github.com/nichurocks02/DCCPPA.git}
    \item cd DCCPPA
    \item follow the readme from here to run the scripts. 
\end{itemize}

\subsection{Scenarios}
For achieving the goal of this research paper which is to compare the DCCPPA algorithm with PRM and RRT, and to show how efficient DCCPPA algorithm is, we consider the following scenario for the experiment.
\begin{itemize}
    \item \textbf{Scenario:} Multiple run for same set of obstacles for different algorithms.
\end{itemize}
\hfill\break
Let's start with using the same obstacles and configure space and running the path planning from start point to end goal multiple times to find out the performances of the algorithm. 
In the below figure you can see the configure space with the obstacles in it.

\begin{figure}[h!]
    \centering
    \includegraphics[scale=0.2]{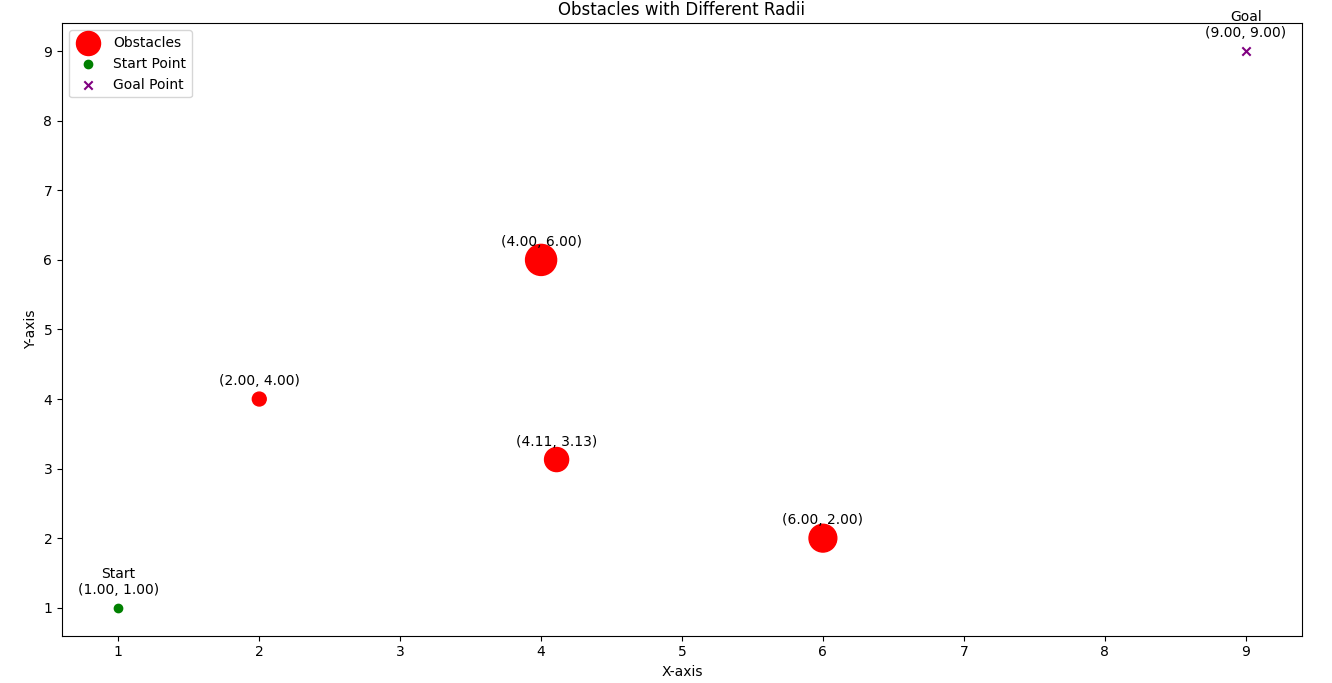}
    \caption{Obstacles with start and goal point for scenario 1}
    \label{fig:my_label}
\end{figure}

\begin{table}[h]
    \centering
    \caption{Comparison of DCCPPA, PRM, and RRT for scenario I}
    \begin{tabular}{|c|c|c|c|}
        \hline
        \textbf{Trial Number} & \textbf{DCCPPA} & \textbf{PRM} & \textbf{RRT} \\
        \hline
        1 & 47 & 204 & 37 \\
        \hline
        2 & 28 & 321 & 53 \\
        \hline
        3 & 23 & 318 & 37 \\
        \hline
        4 & 78 & 184 & 44 \\
        \hline
        5 & 50 & 264 & 88 \\
        \hline
        6 & 74 & 261 & 40 \\
        \hline
        7 & 33 & 417 & 50 \\
        \hline
        8 & 60 & 287 & 55 \\
        \hline
        9 & 44 & 293 & 49 \\
        \hline
        10 & 43 & 294 & 78 \\
        \hline
    \end{tabular}
\end{table}

\textbf{NOTE:} \textit{Each row in the above table describes the no of nodes plotted to reach the goal point from the start point for each algorithm.}

Lets analyze the results from the above table. As we can see that each row represents the number of nodes taken from start point to end goal without colliding with the obstacles, we observe that probabilistic random map takes the most number of steps or rather builds a graph with most number of nodes to reach the end goal, i.e. it takes on a average of approx 284 steps where as the average of rapidly random tree is of approx 53 steps and lastly the dccppa algorithm takes an average of 48 steps. This clearly states that the dynamic curvature constrained path planning algorithm is on par with rrt and far better than prm. The reader also needs to take into account that we have simulated the path generation only for 10 trials for fixed obstacle map, one can easily reconstruct the experiment and analyze for more trials with different obstacle space map, it can be part of future work and is out of scope for this paper.

Below you can see the constructed map from start to goal point for dccppa from one of the above trial. 
\begin{figure}[h!]
    \centering
    \includegraphics[scale=0.2]{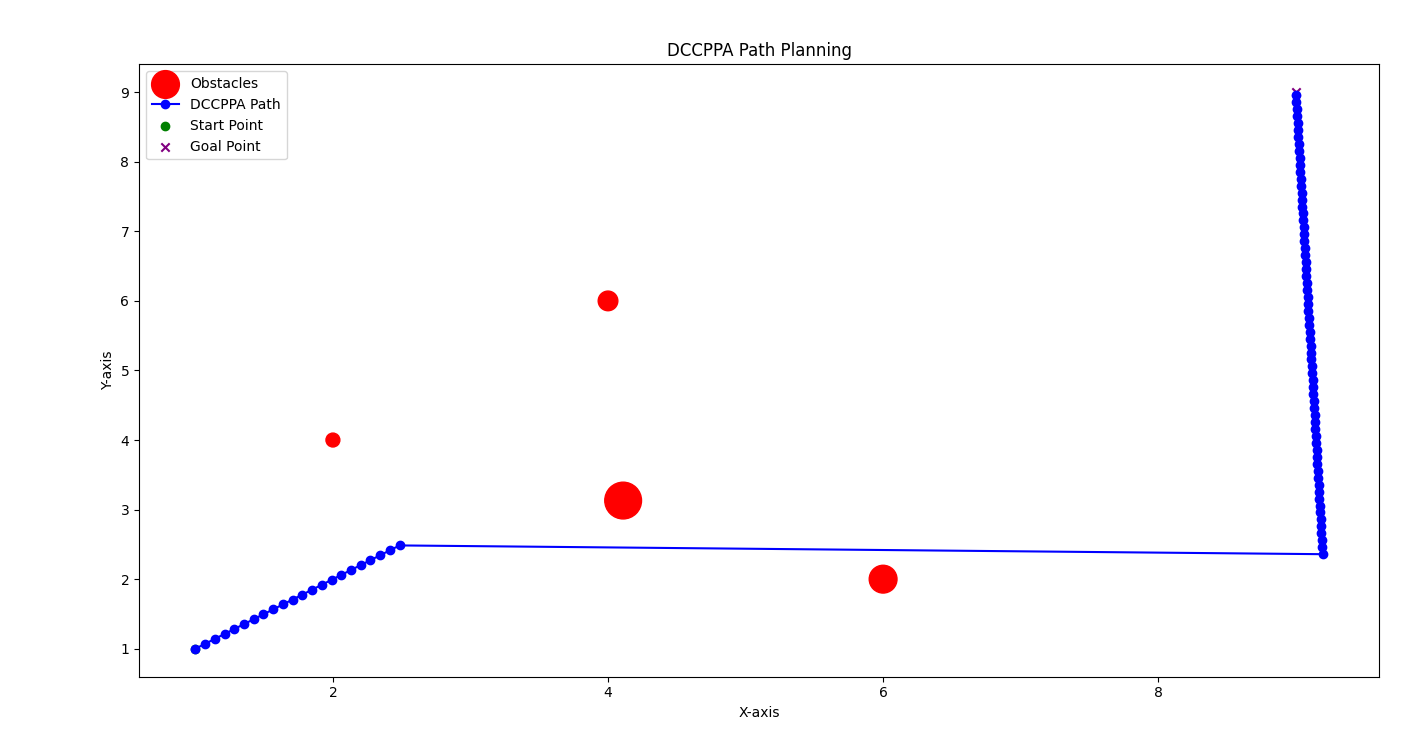}
    \caption{Path construction of DCCPPA algorithm}
    \label{fig:my_label}
\end{figure}

\hfill\break

\subsection{Code Complexity of DCCPPA algorithm}
The code complexity of an algorithm provides insights into how the computational requirements grow as the input size increases. Analyzing the code complexity of DCCPPA involves assessing its fundamental operations and how they scale with varying input parameters. In the case of DCCPPA, we can consider the following aspects:

\subsubsection{Time Complexity}
The time complexity of DCCPPA is influenced by several factors:
\begin{itemize}
    \item Global Search Iterations: The number of iterations in the global search loop affects the overall time complexity. As the number of iterations increases, so does the time complexity.
    \item Local Search Steps: The steps taken in the local search process contribute to the time complexity. The complexity depends on the distance the algorithm needs to traverse in the local search space.
    \item Path Planning Loop: The main loop for path planning, governed by the convergence criteria, also influences time complexity.
\end{itemize}

The overall time complexity of DCCPPA is a combination of these factors and can be expressed as O(N * M), where N is the number of global search iterations, and M is the number of steps in the local search or path planning loop.

\subsubsection{Space Complexity}
The space complexity of DCCPPA involves the memory requirements during its execution:
\begin{itemize}
    \item Vertices and Path Storage: The space required to store vertices and the resulting path contributes to space complexity. This is influenced by the number of iterations and the length of the final path.
    \item Obstacles: The memory required to store obstacle information is constant and depends on the number of obstacles.
\end{itemize}

The space complexity of DCCPPA can be expressed as O(N + M), where N is the space required for storing vertices and path, and M is the space required for obstacle information.
In conclusion, while DCCPPA provides a unique approach to path planning, its time complexity is influenced by the number of iterations and steps in the local search, and the space complexity is driven by the storage requirements for vertices, path, and obstacles. The algorithm's design allows for flexibility and adaptability in different problem domains.

This analysis provides a high-level overview of the code complexity aspects of the DCCPPA algorithm, aiding developers and researchers in understanding its computational characteristics.

\section{Conclusion and Future Work}
\subsection{Conclusion}
In this study, we derived a new path planning algorithm called dynamic curvature constrained path planning algorithm and investigated the performance of it w.r.t algorithms: Probabilistic Road Map (PRM), and Rapidly Exploring Random Tree (RRT). Our experiments, as detailed in the results section, revealed valuable insights into the strengths and limitations of each algorithm.

DCCPPA demonstrated a very high efficiency in path planning along with it's adaptability to curvature constraints, and according to the results being the algorithm that has taken the least number of steps to reach the goal from the start point, sometimes equally or even better than Rapidly Exploring Random Tree algorithm and far better than Probabilistic Road Map. It portrayed consistent performance in multiple trials. The algorithm's ability to dynamically adapt to various obstacles showcases its potential for real-world applications where path planning plays a vital role.

While PRM and RRT exhibited distributed exploration, randomized growth and had a very good exploration efficiency, they also faced challenges in path length variability, sub optimal paths and were sensitive to parameters. These findings contribute to the ongoing discourse on efficient path planning strategies.

\subsection{Future Work}
Our study opens avenues for future research in several directions. Starting with the question on how the DCCPPA algorithm performs in real life scenarios, it's ability to adapt not just to 2D spaces but also 3D spaces, does the algorithm's efficiency increases or decreases if the constraints are increased/decreased , does the space boundary matter in finding the optimal path, chances of DCCPPA being used along with other path planning algorithms, performance of DCCPPA in case of dynamic movement of constraints, it's support for multi-agent systems in 3d spaces, exploring optimization techniques to further improve the efficiency, conducting extensive benchmarking against other state-of-the-art algorithms in diverse scenarios to establish DCCPPA's strengths and weaknesses more comprehensively, the possibility of incorporating Machine Learning techniques to it and last but not the least developing visualization tools to help users understand how the algorithm plans paths and make its decision, enhancing transparency and trust in its behavior. 
These future work directions aim to enhance the capabilities, robustness, and applicability of the DCCPPA algorithm in a variety of scenarios and domains.

\end{document}